%% file: neurips_2025.tex
\documentclass{article}

    \usepackage[preprint]{neurips_2025}

\usepackage[utf8]{inputenc} 
\usepackage[T1]{fontenc}    
\usepackage{hyperref}       
\usepackage{url}            
\usepackage{booktabs}       
\usepackage{amsfonts}       
\usepackage{nicefrac}       
\usepackage{microtype}      
\usepackage{xcolor}         
\usepackage{amsmath}
\usepackage{amssymb}
\usepackage{mathtools}
\usepackage{amsthm}
\usepackage{microtype}
\usepackage{graphicx}
\usepackage{subfigure}
\usepackage{wrapfig}

\usepackage{bbm}
\usepackage{colortbl}
\usepackage{float}
\usepackage{multirow}
\usepackage{multicol}

\usepackage[normalem]{ulem}
\useunder{\uline}{\ul}{}
\usepackage[capitalize,noabbrev]{cleveref}
\theoremstyle{plain}
\newtheorem{theorem}{Theorem}[section]
\newtheorem{proposition}[theorem]{Proposition}

\theoremstyle{definition}
\newtheorem{definition}[theorem]{Definition}

\theoremstyle{remark}

\DeclareMathOperator*{\argmin}{arg\,min}

\title{Minimal Semantic Sufficiency \\Meets Unsupervised Domain Generalization}

\author{%
  Tan Pan$^{1,2}$\thanks{\* Equal contribution.}, ~Kaiyu Guo$^{3,2}$\footnotemark[1],~Dongli Xu$^2$,~Zhaorui Tan$^2$,~Chen Jiang$^{1,2}$\thanks{\* Corresponding authors. This research was conducted during an internship at the Shanghai Academy of Artificial Intelligence for Science.},~Deshu Chen$^{1}$\\
  \textbf{~Xin Guo$^{1,2}$,~Brian C. Lovell$^{3}$,~Limei Han$^{1,2}$,~Yuan Cheng$^{1,2}$\footnotemark[2], Mahsa Baktashmotlagh$^{3}$} \\
  \textsuperscript{1} AI$^3$, Fudan University \textsuperscript{2} Shanghai Academy of Artificial Intelligence for Science \\
\textsuperscript{3} The University of Queensland \\
{\tt\small {pant23@m.fudan.edu.cn, jiangchen@sais.org.cn, cheng\_yuan@fudan.edu.cn}}}

\begin{document}
\maketitle

\begin{abstract}

The generalization ability of deep learning has been extensively studied in supervised settings, yet it remains less explored in unsupervised scenarios. Recently, the Unsupervised Domain Generalization (UDG) task has been proposed to enhance the generalization of models trained with prevalent unsupervised learning techniques, such as Self-Supervised Learning (SSL). UDG confronts the challenge of distinguishing semantics from variations without category labels. Although some recent methods have employed domain labels to tackle this issue, such domain labels are often unavailable in real-world contexts. In this paper, we address these limitations by formalizing UDG as the task of learning a \textbf{M}inimal \textbf{S}ufficient \textbf{S}emantic \textbf{R}epresentation: a representation that (i) preserves all semantic information shared across augmented views (sufficiency), and (ii) maximally removes information irrelevant to semantics (minimality). We theoretically ground these objectives from the perspective of information theory, demonstrating that optimizing representations to achieve sufficiency and minimality directly reduces out-of-distribution risk. Practically, we implement this optimization through Minimal-Sufficient UDG (MS-UDG), a learnable model by integrating (a) an InfoNCE-based objective to achieve sufficiency; (b) two complementary components to promote minimality: a novel semantic-variation disentanglement loss and a reconstruction-based mechanism for capturing adequate variation. Empirically, MS-UDG sets a new state-of-the-art on popular unsupervised domain-generalization benchmarks, consistently outperforming existing SSL and UDG methods, without category or domain labels during representation learning.

\end{abstract}

\input{chapters/introduction}

\input{chapters/relatedwork}

\input{chapters/problemformulation}
\input{chapters/algorithm}
\input{chapters/experiments}

\bibliographystyle{abbrv}
\bibliography{example_paper}

\end{document}

%% file: chapters/introduction.tex
\section{Introduction}
\label{intro}

Generalization ability is a critical yet challenging problem in deep learning.  This challenge leads to the emergence of the task of Domain Generalization (DG)~\cite{zhang2022towardsdisentagle,yu2024insure,tan2024interpret}, which focuses on generalizing deep learning models to unseen distributions. Many works have been proposed on this topic under the supervised settings~\cite{wang2022generalizingdgsurvey,liu2025medmap}. However, for the scenario of unsupervised learning, \textit{e.g.,} self-supervised learning (SSL), which is more prevalent and practical in the real world~\cite{oquab2023dinov2,gui2024surveyssl,tan2024personalize}, limited works have been proposed to address the generalization ability of the model. Recently, the task of Unsupervised Domain Generalization (UDG)~\cite{zhang2022towards} has been proposed to handle the generalization issue in self-supervised learning (SSL), and some methods have been proposed to improve the generalization ability by learning or disentangling the domain-invariant representations~\cite{zhang2025disentangling, zhang2022towards,yang2022cycle}

However, two key challenges remain. (i) Traditional DG methods~\cite{niu2015multimvdg,nam2021reducingsagnet} often focus on disentangling semantics from variation factors~\cite{zhang2022towardsdisentagle}, which becomes more challenging in UDG due to the absence of category labels (\textit{e.g.}, cat, dog, and airplane). (ii) In order to disentangle meaningful semantics from domain-specific variations, existing UDG techniques predominantly depend on domain labels (\textit{e.g.}, painting, scratch, and clipart) that are inaccessible or expensive in real-world scenarios.  Motivated by these challenges, our work addresses a critical research question: \textit{How can semantic information be effectively disentangled in UDG without relying on domain labels?}



To address this research question, we explore learning semantics from an information-theoretical  perspective~\cite{alemi2016deep,wang2020understanding}. Specifically, we argue that the shared information learned by contrastive learning is adequate to represent semantics~\cite{tsai2021selfsupervisedminimal}. However, it may also encompass semantically irrelevant but confounding factors, such as shared style and texture, which can lead to suboptimal semantic representations. This issue might be exacerbated when unlabeled data exhibit significant semantic and covariate distribution shifts. Therefore, eliminating semantically irrelevant information is crucial to refining sufficient semantic representations for UDG.

Based on the above assumption, our intuition is to separate the semantic and variation information by minimizing semantically irrelevant information within sufficient representations. To achieve this, we formulate this non-trivial challenge as a constrained optimization problem, aiming to reduce the dependency between the sufficient representation and the input conditioned on the semantic space. Concretely, the optimization problem can be decomposed into two objectives: (1) minimizing the mutual information between the semantic and variation representations, and (2) maximizing the mutual information between the variations and the inputs conditioned on the semantic space. Objective 1 ensures that semantics and variations are disentangled, promoting their independence and non-redundancy. Objective 2 enables the semantic representation to retain minimal semantically irrelevant information. By optimizing these two objectives, the system is able to learn approximately optimal semantics that is both disentangled and minimally sufficient.

To achieve these objectives, we introduce a novel algorithm, namely Minimal Sufficient UDG (MS-UDG). The model first employs traditional contrastive learning to ensure a sufficient semantic representation. Then, two modules, \textit{i.e.}, Information Disentanglement Module (IDM) and Semantic Representation Optimizing Module (SROM), are employed to disentangle the optimal semantic representation from the sufficient semantic representation. IDM separates the semantic and variation representations from the sufficient semantic representation. SROM applies a mixed InfoNCE constraint to minimize the mutual information between semantics and variations. Simultaneously, it maximizes the mutual information between variations and the inputs through reconstruction, thereby fulfilling our learning objectives. 

\textbf{Contributions.} 
(1) To the best of our knowledge, we propose the first theoretical optimal semantic representation for UDG from an SSL perspective. Subsequently, we introduce a tractable estimation method for disentangling the optimal semantic representation through two optimization objectives. This approach offers a novel view for enhancing the generalization capabilities of SSL models to previously unseen domains. 

(2) Based on dual optimization objectives, we introduce MS-UDG, an algorithm that effectively removes semantically irrelevant information while preserving representative semantics, without relying on domain labels. We also provide theoretical analysis to support the rationale and applicability of our framework.

(3) Experimental results on popular benchmark UDG datasets demonstrate that our method achieves superior performance in downstream tasks compared to existing approaches.

%% file: chapters/relatedwork.tex
\section{Related Work}
\textbf{Unsupervised Domain Generalization.}
Unsupervised Domain Generalization (UDG) has been proposed to handle the problem of domain generalization in unsupervised learning~\cite{zhang2022towards}.
Similar to supervised domain generalization~\cite{nam2021reducing,mitrovic2021representation, hu2020domain, zhang2022towardsdisentagle, guo2024domain}, methods in UDG mostly rely on representation learning and data augmentation to conduct domain-invariant or domain-specific self-supervised learning (SSL). For data augmentation, FDA~\cite{xu2021fourier} and BSS~\cite{scalberttowards} introduce Fourier-based methods to standardize the style of images to plug in native contrastive-based SSL. In addition, BrAD\cite{harary2022unsupervised} generalizes the model by aligning the image to a unified style on the feature level. For representation learning, contrastive-based methods construct negative samples~\cite{zhang2022towards} or suppress intra-domain connectivity by domain labels~\cite{liu2023promoting}. Also, MAE~\cite{he2022masked} is utilized in UDG. 
DiMAE~\cite{yang2022domain} and CycleMAE~\cite{yang2022cycle} transform the original image into its style-mixed view and then decode different domain styles by several domain decoders. DisMAE~\cite{zhang2025disentangling} introduces a semantic-encoder and a variation-encoder to disentangle semantic attributes. Moreover, some methods use UDG methods to improve the generalization on specific tasks, such as face anti-spoofing~\cite{liu2023towards}. Unlike these methods, we theoretically build our method from the contrastive learning perspective.\par
\textbf{Contrastive Learning.}
Contrastive learning~\cite{oquab2023dinov2,chen2020simple,pan2025structure} is a successful paradigm in SSL. Contrastive-based SSL methods aim to learn the shared information~\cite{oord2018representation} between multi-view data. Recent methods~\cite{wang2022rethinkingminimal,tsai2021selfsupervisedminimal,liang2024factorized} discuss task-relevant and redundant representations of SSL based on information theory. Although there has been extensive exploration in the SSL field, the generalization of SSL to OOD data still requires further discussion.

\textbf{Disentangled Representation Learning.}
Disentangled representation learning is a promising approach for DG, aiming to separate semantics-relevant and irrelevant factors in data~\cite{wang2024disentangled,zhang2022towards,kim2018disentangling,zhou2022domain,bui2021exploiting，}. Class labels help ensure the model learns class-specific features consistently across domains, avoiding reliance on domain-specific artifacts. Without class labels, the disentanglement-based methods in UDG often focus on domain style transfer~\cite{yu2024insure,zhang2022towardsdisentagle} to separate domain variations by domain labels. However, there is still a lack of theoretical support for disentangling domain-invariant representations of SSL to obtain optimal semantics.

%% file: chapters/problemformulation.tex
\section{Preliminary and Motivation}
In Sec. \ref{sec:3.1}, we first introduce the setting of UDG, which aims to discard the covariate information in the representations~\cite{zhang2025disentangling}. To address our research question, in Sec.~\ref{sec:3.2}, we propose a new concept of minimal sufficient semantic representation, treating semantic information as a proxy for the unknown downstream task. Based on this concept, we theoretically formulate two learning objectives optimized to disentangle representation into pure semantic information. 
Finally, based on these objectives, we present our algorithm and provide a theoretical analysis in Sec.~\ref{algorithm}.
\subsection{Unsupervised Domain Generalization}
\label{sec:3.1}
\textbf{Notation.} Let $\mathcal{X}$ be the input space and $\mathcal{Y}$ be the label space. Considering a supervised dataset $D=(X_D,Y_D)$ with $N_D$ samples, where $X_D\subset\mathcal{X}$ and $Y_D\subset\mathcal{Y}$, the distribution of $D$ is $P_{(X_D, Y_D)}$. Then we can define the distribution of SSL data, downstream supervised data, and test data as $P_{(X_{ssl},\emptyset)}$, $P_{(X_{sup},Y_{sup})}$ and $P_{(X_{test},Y_{test})}$ respectively. 
In SSL, for each input $(x_1, x_2)$, we have $x_1\in X_{ssl}$ and $x_2\in X_{aug}$, where $X_{aug}$ is augmented from $X_{ssl}$. The extracted features are $z_1=h(x_1)$ and $z_2=h(x_2)$, where $h$ is the encoder and $z_1,z_2\in \mathcal{Z}$, $\mathcal{Z}$ is the representation space. The parameterized latent space for semantic-relevant and -irrelevant, or variate, factors are $\mathcal{S}$ and $\mathcal{V}$, respectively. The intuitive relation is $\mathcal{Z} = \mathcal{S}\oplus\mathcal{V}$.   \par
\textbf{Problem setting.} Following the setting of UDG~\cite{zhang2022towards}, the test distribution should be kept unknown at any stage before inference. Then, we have the condition $Support(P(X_{test}))\cap(Support(P(X_{sup}))\cup Support(P(X_{ssl})))=\emptyset$. In addition, to avoid the semantic shift in the downstream task, we need $Support(P(Y_{sup})) = Support(P(Y_{test}))$. The target of UDG is to obtain the optimized feature extractor $h^*$ which has the minimum risk on the test distribution $h^* =\argmin_h R_{P_{(X_{test},Y_{test})}}(h)$

\textbf{Motivation.} Considering that UDG methods are built upon existing SSL methods with unknown downstream task targets $T$, we examine the goal of UDG from a contrastive learning (CL) perspective. Given an input $x_1$ and its augmented counterpart $x_2$, we adhere to the multi-view assumption~\cite{tsai2021selfsupervisedminimal}, treating $x_1$ and $x_2$ as two corresponding views of the same data point. Under this assumption, the objective of CL is to extract task-relevant representations~\cite{tsai2021selfsupervisedminimal}. \cite{tsai2021selfsupervisedminimal} posits a compression gap $I(x_1;x_2\big\vert T)$ under the task $T$. This implies that while optimizing for $I(x_1;x_2)$, task-irrelevant information may inadvertently be retained in the learned representations. In datasets used for SSL, especially those from multiple domains that exhibit significant semantic and covariate distribution shifts, this compression gap cannot be overlooked if the aim is to achieve task-specific or semantic-specific representations. Based on this insight, we propose that semantics and variations can be separated by explicitly modeling and minimizing task-irrelevant information.

\subsection{Minimal Sufficient Semantic Representation}
\label{sec:3.2}

In this section, we first introduce the definition of sufficient representation and minimal sufficient representation~\cite{wang2022rethinkingminimal,tsai2021selfsupervisedminimal}, which aims to learn task-relevant representation optimally.
In the supervised setting, with the given image $x$ and label $y$, the information bottleneck~\cite{tishby2000informationbottleneck} can be applied to achieve optimal representations by minimizing mutual information $I(x,z)$ and maximizing $I(z,y)$, where $z$ is the representation of $x$.  In self-supervised learning, the augmented image $x_2$ plays a similar role as the supervised label~\cite{Federici2020Learninginfobottleneck, tsai2021selfsupervisedminimal}. The definition is proposed as follows.


\begin{definition}[{Minimal Sufficient Representation in Contrastive Learning~\cite{wang2022rethinkingminimal}}]
Let $\hat{z}_1^{suf}$ and $\hat{z}_1^{min}$ be the sufficient representation and minimal sufficient representation of $x_1$ in self-supervised learning, respectively.  $x_2$ is the augmented data. Then the following conditions should be satisfied.   
{\[I(\hat{z}_1^{suf}; x_2) = I(x_1;x_2), \hat{z}_1^{min} = \argmin_{\hat{z}_1^{suf}}I(\hat{z}_1^{suf};x_1)\]}
\label{def:minimalsufficient}
\end{definition}\vspace{-3ex}
This definition introduce the minimal sufficient representation with two stage: First, we define that $\hat{z}_1^{suf}$ is sufficient if and only if $I(\hat{z}_1^{suf}; x_2) = I(x_1;x_2)$, which aligns with the target of contrastive loss to maximize $I(z_1;z_2)$~\cite{oord2018representation}. Note that, since $\hat{z}_1^{suf}$ is defined to keep all the mutual information between $x_1$ and $x_2$ instead of any prior labels, the loss or change of augmented information in $x_2$ can also affect $\hat{z}_1^{suf}$. Second, with $\hat{z}_1^{suf}$, we define the minimal sufficient representation $\hat{z}_1^{min}$ by searching minimum $I(\hat{z}_1^{suf};x_1)$.
Symmetrically, we can also define the minimal sufficient representation of $x_2$. More details of minimal sufficient representation can be found in Appendix~\ref{explainconcepts}

However, the minimal sufficient representation in Definition \ref{def:minimalsufficient} may still include task-irrelevant information, particularly in UDG datasets with significant covariate shifts.
\begin{proposition}
$I(\hat{z}_1^{min};x_i )=I(\hat{z}_1^{min}; T) + I(\hat{z}_1^{min};x_i\big\vert T) \geq I(\hat{z}_1^{min}; T)$ , where $i\in\{1,2\}$, $T$ is the downstream task. 
\label{pro1:assum}
\end{proposition}
%
The proposition \ref{pro1:assum} indicates that if $I(x_i; \hat{z}_1^{min}\big\vert T)$ is large, $\hat{z}_1^{min}$ may not be the optimal minimal sufficient feature. In practice, $I(x_i; \hat{z}_1^{min}\big\vert T)$ can be affected by many factors in the shared information, including style, texture, and other factors between $x_2$ and $x_1$. In the multi-domain situation as the setting of UDG, the $I(x_i; \hat{z}_1^{min}\big\vert T)$ will be large, which leads to performance degradation since there can be significant distribution shifts in all task-irrelevant distributions. \par
The downstream task remains unseen during SSL. However, in most cases, it should be related to the semantic information in $X_{sup}$. In this case, we can consider $I(x_1;x_2; T)=I(x_1;x_2;S)$. Then, we argue that the sufficient representation should contain all semantic information between $x_1$ and $x_2$.

\begin{definition}[Sufficient Semantic Representation in Contrastive Learning]
    $z_1^{suf}$ is the disentangled sufficient representation of $x_1$ if and only if $I(z_1^{suf};x_2;S) = I(x_1;x_2;S)$. $S\subset\mathcal{S}$ is the semantic information in $X_{ssl}$
\label{def:sufficientsemantic}
\end{definition}
Definition \ref{def:sufficientsemantic} indicates that a representation containing all the shared semantic information between $x_1$ and $x_2$ is sufficient to capture the semantics of $x_1$ and $x_2$. However, the sufficient representation may not be the optimal representation as $I(z^{suf};x\big\vert S)\geq 0$, where $I(z^{suf};x)>0$. 

From the relation $\mathcal{Z} = \mathcal{S}\oplus\mathcal{V}$, every $z^{suf}$ can be decomposed as $z^{suf} = (s,v)$, where $s$ is semantic-relevant, which is also a sufficient semantic representation, and $v$ is semantic-irrelevant. That means $I(v;S)=0$. By this disentanglement, if $I(s;x\big\vert S)$ can be minimized, then $s$ is the optimal sufficient representation we expect. From this perspective, we propose the definition of the optimal sufficient semantic representation in Definition~\ref{def:minisuffdis}. 
\begin{definition}[{Minimal Sufficient Semantic Representation in Contrastive Learning}]
     Let $z_1^{min}$ be the minimal sufficient semantic representation of $x_1$. Then, $z_1^{min}  \triangleq \argmin_{z_1^{suf}} I(z_1^{suf};x_1, x_2\big\vert S)$, $S\subset\mathcal{S}$
    is the semantic information in $X_{ssl}$, $x_2$ is the augmented input.
\label{def:minisuffdis}
\end{definition}
Definition~\ref{def:minisuffdis} suggests that the minimal sufficient semantic representation should contain the least semantic-irrelevant information. Furthermore, we can assume $I(z_1^{min};x_1,x_2\big\vert S)=0$. As mentioned above,  we have the decomposition of sufficient semantic representations $\forall z^{suf}, z^{suf}=(z^{min},v)$. Then, we theoretically illustrate how to effectively disentangle the $z^{min}$, which answers the research question we proposed.
\begin{proposition}
    $I(z_1^{suf};x_1\big\vert S) = I(z_1^{suf};x_1, x_2\big\vert S)$
    \label{pro2}
\end{proposition}
Proposition \ref{pro2} indicates that the semantic-irrelevant information in $z_1^{suf}$ is derived from $x_1$, and we can obtain $z_1^{min}$ as $z_1^{min}  \triangleq \argmin_{z_1^{suf}} I(z_1^{suf};x_1\big\vert S)$, 
\begin{proposition}
$z = (s, v)$ is the sufficient semantic representation of $x$ and $s$ is semantic-relevant representation. Then $I(s;x\big\vert S) = I(s;v)+I(z;x\big\vert S)-I(v;x\big\vert S)$
\label{pro3}
\end{proposition}
If we want to optimize $s$ as the minimal sufficient semantic representation, considering the Markov chain $x\rightarrow z\rightarrow s, v$, we can only optimize $s$ and $v$. \textit{\textbf{So, we propose to  minimize $I(s;v)$ and maximize $I(v;x\big\vert S)$ to disentangle the optimal semantic representation in UDG.}}

%% file: chapters/algorithm.tex
\section{Algorithm and Theory}
\label{algorithm}
\subsection{Algorithm}
Based on the objectives in Proposition \ref{pro3}: (1) minimize $I(s;v)$ and (2) maximize $I(v;x\big\vert S)$, we propose that objective 1 enhances the independence and non-redundancy between semantic representation and variations, while objective 2 limits the inclusion of semantic-irrelevant information in semantic representation. 

\begin{figure*}[!htp]
\begin{center}
\centerline{\includegraphics[width=1.0\textwidth]{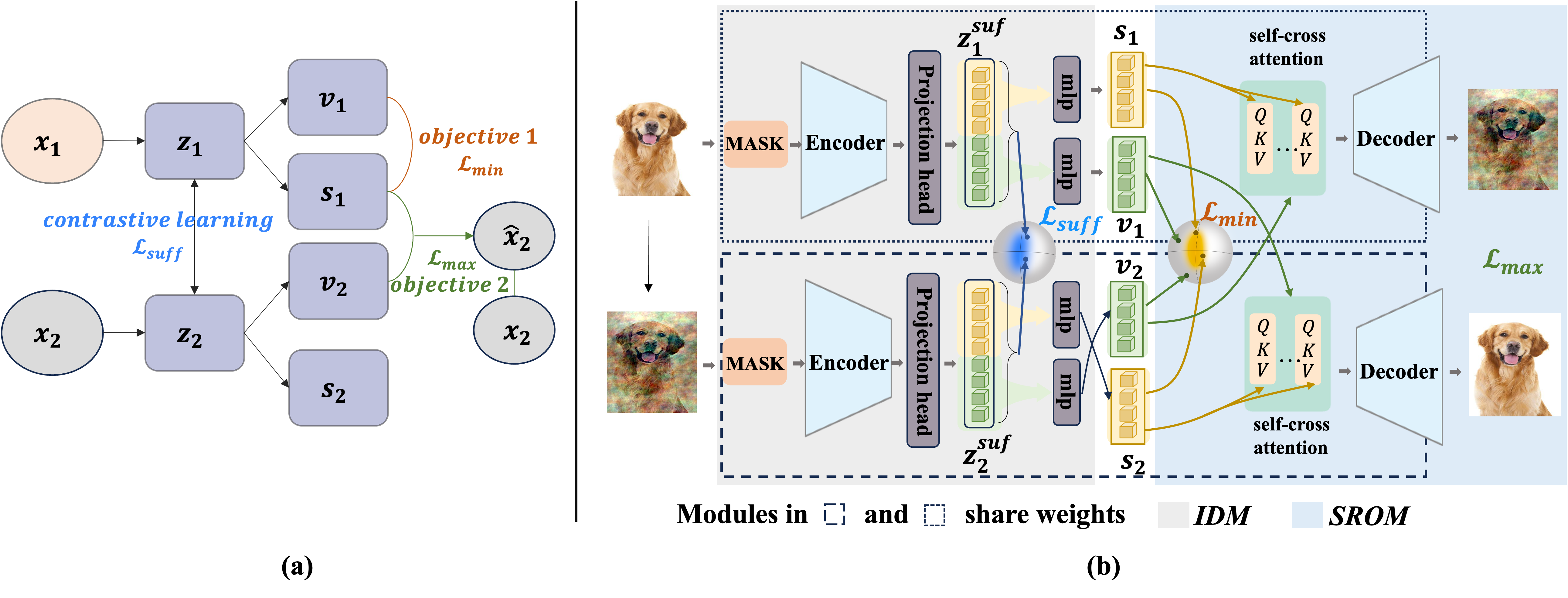}}
\caption{Our method's pipeline, where $x_1$ and $x_2$ denote the input and its augmented counterpart. $z$ represents the sufficient representation, while $s$ and $v$ correspond to the semantic representation and variation. (a) An illustration of our two objectives. First, we employ $\mathcal{L}_{suff}$ to capture sufficient semantics. Subsequently, $\mathcal{L}_{min}$ and $\mathcal{L}_{max}$ are introduced to learn a minimal sufficient representation. We only describe $\mathcal{L}_{min}$ and $\mathcal{L}_{max}$ for $x_1$ in detail. The same procedure applies to $x_2$.  (b) A detailed overview of our network and the applied constraints. }
\label{pipeline}
\end{center}
\vskip -0.2in
\end{figure*}

To achieve objectives, we propose the method Minimal Sufficient UDG (MS-UDG) as shown in Fig.~\ref{pipeline}. After learning sufficient representations by contrastive learning loss InfoNCE~\cite{he2020momentum}, the model comprises two components: (1) Information Disentanglement Module (IDM), which aims to disentangle semantic representation $s$ and variation $v$. (2) Semantic Representation Optimizing Module (SROM), which optimizes $s$ towards minimal sufficient semantic representation.

Firstly, the InfoNCE loss is applied to constrain $z^{suff}$, which aims to learn sufficient shared information between $x_1$ and $x_2$.
\begin{equation}
    \mathcal{L}_{suf}=-\log\frac{exp(z_1^{suff}\cdot z_2^{suff}/\tau)}{\sum_{i=0}^{K}exp(z_1^{suff}\cdot z_k^{suff-}/\tau)},
\end{equation}
where $\tau$ is a temperature hyper-parameter and the K is the number of negative samples. $z_k^{suff-}$ represents a negative representation from the negative sample. We construct positive pairs by Fourier-based augmentations (FA)~\cite{xu2021fourier}, whose efficiency has been experimentally proved in UDG methods~\cite{scalbert2023towards,yang2022cycle}. FA randomly alters the amplitudes of one image (anchor) with the amplitudes from another image (target). For our training, the target image is randomly selected from the dataset.

Then, based on the sufficient semantic representation, we introduce two components.

\textbf{IDM.} Following intuitive relation $\mathcal{Z} = \mathcal{S}\oplus\mathcal{V}$, we assume that the representation $z$ can be disentangled into a semantic-relevant representation $s$ and a semantic-irrelevant representation $v$. After a shared encoder extracts image features, the features are processed through a projection head to obtain a high-dimensional representation. Subsequently, two MLP modules are applied to $z$ to separate semantic information from variation.

\textbf{SROM.} Corresponding to Proposition~\ref{pro3}, we introduce a semantics and variation mixed InfoNCE $L_{min}$ loss to minimize $I(s;v)$ and a translated reconstruction loss $L_{max}$  to maximize $I(v;x\big\vert S)$.

\begin{equation}
\begin{aligned}
    &\mathcal{L}_{min}\!=\!
    -\log\!\frac{\exp(s_1\cdot s_2)}{\sum\limits_{k=0}^{K}\!\exp(s_1\!\cdot\! s_k^-)\!+\!\sum\limits_{k=0}^{K}\!\exp(s_1\!\cdot\! v_k^-)\!+\!\exp(s_1\!\cdot\! v_1)},
\end{aligned}
\end{equation}
where $v_1$ is considered as a negative representation. $\mathcal{L}_{min}$ is modified from InfoNCE loss, which not only maximizes the  $I(s_1;s_2)$ but also minimizes $I(s_1;v_1)$.  
\begin{proposition}
    $\mathcal{L}_{min} \geq I(s_1;v_1)- I(s_1;s_2)$
\label{pro4:min}
\end{proposition}
According to Proposition \ref{pro4:min},  minimizing $\mathcal{L}_{min}$ indicates decreasing $I(s_1;v_1)$ and increasing $I(s_1;s_2)$ synchronously. Minimizing $\mathcal{L}_{min}$ satisfies our target to minimize $I(s;v)$. Meanwhile, same as InfoNCE, $\mathcal{L}_{min}$ also keeps $s_1$ as a sufficient semantic representation by maximizing $I(s_1;s_2)$.  
\begin{equation}
    \mathcal{L}_{max}=\frac{\|\mathbf{D}(s_1, v_2)- x_2 \|_2^2 + \|\mathbf{D}(s_2, v_1)- x_1 \|_2^2}{2}.
\end{equation}
Here, $\mathbf{D}$ is the decoder to reconstruct images.
Without loss of generality, we consider the object $\mathbf{D}(s_1, v_2)$ to illustrate how $\mathcal{L}_{max}$ maximize $I(v;x\big\vert S)$ in the following.
\begin{proposition}
    $I(v_2,s_1;x_2)-I(x_1;x_2)\leq I(v_2;x_2\big\vert S)$
\label{pro4:max}
\end{proposition}
Proposition \ref{pro4:max} indicates that $I(v_2,s_1;x_2)$ is related to the lower-bound of $I(v_2;x_2\big\vert S)$. From $\|\mathbf{D}(s_1, v_2)- x_2 \|_2^2$, $\mathcal{L}_{max}$ aims to maximize the mutual information $I(v_2,s_1;x_2)$. Thus, $\mathcal{L}_{max}$ can help increase the mutual information $I(v_2;x_2\big\vert S)$ by raising its lower bound\par 
As discussed above, $\mathcal{L}_{min}$ and $\mathcal{L}_{{max}}$ meet our learning targets described in Section \ref{sec:3.2}. Therefore, the final loss is
\begin{equation}
\mathcal{L}=\mathcal{L}_{suff}+\mathcal{L}_{min}+\mathcal{L}_{max}.
\end{equation}
\subsection{Theoretical Analysis of Generalization Upper Bound}
 As discussed in previous works~\cite{albuquerque2019generalizing,cha2021swad} on the generalization error bound based on the multi-distribution learning theory~\cite{ben2010theory}, the upper bound of the target risk can be summarized into three parts~\cite{cha2021swad}: source empirical risk, distribution discrepancy between source and target distributions, and confidence bound. Briefly, it can be presented as $R_{P_{(X_{test},Y{test})}}(h)\leq R_{P_{(X_{sup},Y{sup})}}(h) + \Delta\approx e_{sup} +\Delta $,
where $e_{sup}$ is the downstream error, $\Delta$ is the summation of domain discrepancy and confidence bound in downstream~\cite{cha2021swad}. However, we can not access the state of $\Delta$ in the SSL stage. Thus, we focus on the error in the source domain to minimize the generalization bound in the downstream. In the following, we demonstrate the relation between the minimal sufficient semantic representation and the downstream error.
\begin{theorem}
    For representation $z_1^{suf}$ and $z_1^{min}$, their Bayes error rates are $e^{min}$ and $e^{suf}$ respectively. $e^{min}$ has the minimum upper bound compared with all minimal sufficient semantic representations. Specifically, given the downstream task $T$,  we have
    \begin{align*}
     e^{suf}\leq 1- \exp[-(H(T)+I(z_1^{suf};x_1\big\vert S)]\\
     e^{min}\leq 1- \exp[-(H(T)+I(z_1^{min};x_1\big\vert S)]
    \end{align*}
\label{theo}
\end{theorem}\vspace{-3ex}
Bayes error rate~\cite{fukunaga2013introduction} is the minimum achievable error for any representation-learned classifier. Following the previous work~\cite{wang2022rethinkingminimal,tsai2021selfsupervisedminimal}, we consider it as the downstream error. Theorem \ref{theo} indicates that downstream error of sufficient semantic representation $z_1^{suf}$ is upper bounded by $I(z_1^{suf};x_1\big\vert S)$. According to Definition \ref{def:minisuffdis}, the Bayes error rate of $z^{min}$ has the minimum upper bound. Thus, $z^{min}$ provides the lowest upper bound for the risk $R_{P_{(X_{test},Y{test})}}(h)$, which helps to improve the generalization ability of the unsupervised models.

%% file: chapters/experiments.tex
\begin{table*}[!h]
\caption{Performances of UDG and SSL models on PACS dataset. All models are trained on 3 selected domains and tested on the left domains, which process is repeated for all domains. ``avg.'' represents macro-accuracy. We report the accuracy for every domain and the average accuracy for all domains. \textcolor[HTML]{00009B}{\textbf{Best}} and \underline{second best} are highlighted.}
\vskip 0.15in
\centering
\resizebox{0.95\textwidth}{!}{%
\begin{tabular}{l|ccccc|ccccc}
\toprule[1pt]
 & \multicolumn{5}{c|}{Label Fraction: 1\%} & \multicolumn{5}{c}{Label Fraction: 5\%} \\ \cline{2-11} 
 & \multicolumn{5}{c|}{Target domain} & \multicolumn{5}{c}{Target domain} \\
\multirow{-3}{*}{Methods} & \textit{photo} & \textit{art} & \textit{cartoon} & \textit{sketch} & \textbf{\textit{avg.}} & \textit{photo} & \textit{art} & \textit{cartoon} & \textit{sketch} & \textbf{\textit{avg.}} \\ \hline
ERM & 10.90 & 11.21 & 14.33 & 18.83 & 13.82 & 14.15 & 18.67 & 13.37 & 18.34 & 16.13 \\
MoCo V2 & 22.97 & 15.58 & 23.65 & 25.27 & 21.87 & 37.39 & 25.57 & 28.11 & 31.16 & 30.56 \\
SimCLR V2 & 30.94 & 17.43 & 30.16 & 25.20 & 25.93 & 54.67 & 35.92 & 35.31 & 36.84 & 40.68 \\
BYOL & 11.20 & 14.53 & 16.21 & 10.01 & 12.99 & 26.55 & 17.79 & 21.87 & 19.65 & 21.46 \\
AdCo & 26.13 & 17.11 & 22.96 & 23.37 & 22.39 & 37.65 & 28.21 & 28.52 & 30.35 & 31.18 \\
MAE & 30.72 & 23.54 & 20.78 & 24.52 & 24.89 & 32.69 & 24.61 & 27.35 & 30.44 & 28.77 \\
DARLING & 27.78 & 19.82 & 27.51 & 29.54 & 26.16 & 44.61 & 39.25 & 36.41 & 36.53 & 39.20 \\
DiMAE & 48.86 & 31.73 & 25.83 & 32.50 & 34.73 & 50.00 & 41.25 & 34.40 & 38.00 & 40.91 \\
BrAD & {\color[HTML]{00009B} \textbf{61.81}} & 33.57 & 43.47 & 36.37 &  43.80 & \underline{65.22} & 41.35 & 50.88 &  50.68 & 52.03 \\
CycleMAE & \underline{ 52.63} & 36.25 & 35.53 & 34.85 & 39.82 &  63.24 & 39.96 & 42.15 & 36.35 & 45.43 \\
BSS/SimCLR & 43.31 & \underline{38.96} & {\color[HTML]{00009B} \textbf{48.61}} & {\color[HTML]{00009B} \textbf{48.76}} & \underline{44.91} & 58.16 & \underline{46.37} & {\color[HTML]{00009B} \textbf{55.69}} & {\color[HTML]{00009B} \textbf{65.63}} & \underline{56.40}  \\
\rowcolor[HTML]{EFEFEF} 
\textbf{MS-UDG} & 42.74 & {\color[HTML]{00009B} \textbf{47.13}} & \underline{ 47.84} & \underline{ 43.33} & \color[HTML]{00009B} \textbf{45.26} & {\color[HTML]{00009B} \textbf{74.03}} &  {\color[HTML]{00009B}\textbf{61.78}} & \underline{ 54.72} & \underline{ 61.53} & {\color[HTML]{00009B} \textbf{63.02}}\\ \hline
\multicolumn{1}{c|}{} & \multicolumn{5}{c|}{Label Fraction: 10\%} & \multicolumn{5}{c}{Label Fraction: 100\%} \\ \cline{2-11} 
\multicolumn{1}{c|}{} & \multicolumn{5}{c|}{Target Domain} & \multicolumn{5}{c}{Target Domain} \\
\multicolumn{1}{c|}{\multirow{-3}{*}{Methods}} & \textit{photo} & \textit{art} & \textit{cartoon} & \textit{sketch} & \textbf{\textit{avg.}} & \textit{photo} & \textit{art} & \textit{cartoon} & \textit{sketch} & \textbf{\textit{avg.}} \\ \hline
ERM & 16.27 & 16.62 & 18.40 & 12.01 & 15.82 & 43.29 & 24.27 & 32.62 & 20.84 & 30.26 \\
MoCo V2 & 44.19 & 25.85 & 35.53 & 24.97 & 32.64 & 59.86 & 28.58 & 48.89 & 34.79 & 43.03 \\
SimCLR V2 & 54.65 & 37.65 & 46.00 & 28.25 & 41.64 & 67.45 & 43.6 & 54.48 & 34.73 & 50.06 \\
BYOL & 27.01 & 25.94 & 20.98 & 19.69 & 23.40 & 41.42 & 23.73 & 30.02 & 18.78 & 28.49 \\
AdCo & 46.51 & 30.31 & 31.45 & 22.96 & 32.81 & 58.59 & 29.81 & 50.19 & 30.45 & 42.26 \\
MAE & 35.89 & 25.59 & 33.28 & 32.39 & 31.79 & 36.84 & 25.24 & 32.25 & 34.45 & 32.20 \\
DARLING & 53.37 & 39.91 & 46.41 & 30.17 & 42.46 & 68.86 & 41.53 & 56.89 & 37.51 & 51.20 \\
DiMAE & {\color[HTML]{00009B} \textbf{77.87}} & 59.77 & 57.72 & 39.25 & 58.65 & 78.99 & 63.23 & 59.44 & 55.89 & 64.39 \\
BrAD & 72.17 & 44.20 & 50.01 & 55.66 & 55.51 & - & - & - & - & -  \\
CycleMAE & 85.94 &  {\color[HTML]{00009B} \textbf{67.93}} & 59.34 & 38.25 & \underline{ 62.87} & {\color[HTML]{00009B} \textbf{90.72}} & {\color[HTML]{00009B} \textbf{75.43}} & \underline{ 69.33} & 50.24 & \underline{ 71.41} \\
BSS/SimCLR & 63.29 & 51.37 & \underline{59.43} & \underline{66.09} & 60.04 & 79.50 & 62.73 & 65.67 & {\color[HTML]{00009B}\textbf{73.02}} & 70.23  \\
\rowcolor[HTML]{EFEFEF} 
\textbf{MS-UDG} & \underline{74.10} &\underline{ 61.90} & {\color[HTML]{00009B} \textbf{63.73}} & {\color[HTML]{00009B} \textbf{73.66}} & {\color[HTML]{00009B} \textbf{68.35}} & \underline{ 84.80} & \underline{ 63.74} & {\color[HTML]{00009B} \textbf{71.73}} & \underline{ 71.27} & {\color[HTML]{00009B} \textbf{72.89}} \\ 
\bottomrule[1pt]
\end{tabular}%
}
\vskip -0.1in
\label{tab:PACS}
\end{table*}
\section{Experiments}
\label{experiments}
\subsection{Experimental Setup}

\textbf{Datasets.} Following previous UDG work~\cite{zhang2022towards}, PACS~\cite{li2017deeper} and DomainNet~\cite{peng2019moment} are evaluated for benchmarking UDG methods. Meanwhile, we also evaluate our methods on OfficeHome, Office31, and VLCS whose results and details can be found in the Appendix. 

DomainNet has 345 categories and 6 different domains (Clipart, Infograph, Quickdraw, Painting, Real, and Sketch). PACS has 9,991 images with 7 classes from 4 domains: Art, Cartoons, Photos, and Sketches. For DomainNet, the six domains are split into two sets: 1. Clipart, Infograph, and Quickdraw; 2. Painting, Real, and Sketch as previous work did. 20 classes are selected as both unlabeled and labeled data. We use one set for seen domains and the other one for unseen domains. We report the micro-accuracy and macro-accuracy on six domains. For PACS, three domains are selected for training, and the remaining domain is used for evaluation. 

\textbf{Experimental Protocol.} 
We adopt the all-correlated setting, as proposed in DARLING~\cite{zhang2022towards} and DisMAE~\cite{zhang2025disentangling}. The overall process is divided into three main steps. First, a pre-trained model is obtained using UDG methods on the unlabeled data from the source domain. Second, this pre-trained model is fine-tuned on varying proportions of labeled data from the source domain, with either the classifier or the entire backbone network being adjusted. Finally, the model is tested on the unseen domain. Following the benchmark~\cite{zhang2022towards} and general setting of SSL (i.e., linear probing and fine-tuning), we adopt linear probing for 1\% and 5\% label fractions and fine-tuning for 10\% and 100\% label fractions in downstream tasks.


To ensure a fair comparison with previous SSL methods, including MoCo V2~\cite{chen2020improved}, SimCLR V2~\cite{chen2020simple}, BYOL~\cite{chen2020big}, AdCo~\cite{hu2021adco}, and MAE~\cite{he2022masked}, as well as UDG methods such as DARLING~\cite{zhang2022towards}, DiMAE~\cite{yang2022domain}, BrAD~\cite{harary2022unsupervised}, BSS~\cite{scalberttowards}, and CycleMAE~\cite{yang2022cycle}, the SSL models are initialized using ImageNet pre-trained model for the DomainNet and PACS datasets. Additionally, SSL training without ImageNet pre-trained model is conducted for comparison with DisMAE~\cite{zhang2025disentangling} on DomainNet. Further details of this experiment can be found in the Appendix.

\textbf{Implementation Details.} Following previous disentanglement-based UDG methods~\cite{yang2022cycle,yang2022domain}, we adopt ViT-S/16~\cite{dosovitskiy2020image} as the backbone. The learning rate for pre-training is set to $1\times10^{-4}$. The weight decay is set to 0.05, and the batch size is 32. For $\tau$ in $L_{suff}$, we set the temperature $\tau=0.07$ as previous work~\cite{he2020momentum}. During fine-tuning, all methods are trained for 50 epochs, and the best evaluation model is selected to test on unseen domain data, following the exact training schedule outlined in~\cite{zhang2025disentangling}. The test domains also remain unseen during the model selection. Further details can be found in the Appendix.

\begin{table*}[]
\vskip -0.1in
\caption{Performances of UDG and SSL models on the DomainNet subset, which contains six domains. The models are trained on source domains and evaluated on target domains. ``Overall'' and ``avg.'' refer to micro-accuracy and macro-accuracy.  \textcolor[HTML]{00009B}{\textbf{Best}} and \underline{second best} are highlighted.}
\vskip 0.1in
\centering
\resizebox{0.9\textwidth}{!}{%
\begin{tabular}{lcccccccc}
\toprule[1pt]
\multicolumn{9}{c}{Label Fraction: 1\%} \\ \hline
\multicolumn{1}{l|}{Source} & \multicolumn{3}{c|}{\textit{painting$\cup$real$\cup$sketch}} & \multicolumn{3}{c|}{\textit{clipart$\cup$infograph$\cup$quichdraw}} &  &  \\
\multicolumn{1}{l|}{Target} & \multicolumn{3}{c|}{\textit{clipart$\cup$infograph$\cup$quichdraw}} & \multicolumn{3}{c|}{\textit{painting$\cup$real$\cup$sketch}} & \textbf{\textit{overall}} & \textbf{\textit{avg.}} \\ \hline 
\multicolumn{1}{l|}{BYOL} & 6.21 & 3.48 & \multicolumn{1}{c|}{4.27} & 5.00 & 8.47 & \multicolumn{1}{c|}{4.42} & 5.61 & 5.31 \\
\multicolumn{1}{l|}{MoCo V2} & 18.85 & 10.57 & \multicolumn{1}{c|}{6.32} & 11.38 & 14.97 & \multicolumn{1}{c|}{15.28} & 12.12 & 12.9 \\
\multicolumn{1}{l|}{AdCo} & 16.16 & 12.26 & \multicolumn{1}{c|}{5.65} & 11.13 & 16.53 & \multicolumn{1}{c|}{17.19} & 12.47 & 13.15 \\
\multicolumn{1}{l|}{SimCLR V2} & 23.51 & 15.42 & \multicolumn{1}{c|}{5.29} & 20.25 & 17.84 & \multicolumn{1}{c|}{18.85} & 15.46 & 16.86 \\
\multicolumn{1}{l|}{MAE} & 22.38 & 12.62 & \multicolumn{1}{c|}{10.50} & 17.86 & 24.57 & \multicolumn{1}{c|}{19.33} & 17.57 & 17.88 \\
\multicolumn{1}{l|}{DARLING} & 18.53 & 10.62 & \multicolumn{1}{c|}{12.65} & 14.45 & 21.68 & \multicolumn{1}{c|}{21.30} & 16.56 & 16.54 \\
\multicolumn{1}{l|}{DiMAE} & 26.52 & 15.47 & \multicolumn{1}{c|}{15.47} & 20.18 & 30.77 & \multicolumn{1}{c|}{20.03} & 21.85 & 21.41 \\
\multicolumn{1}{l|}{BrAD} & 47.26 & 16.89 & \multicolumn{1}{c|}{{\ul 23.74}} & 20.03 & 25.08 & \multicolumn{1}{c|}{31.67} & 25.85 & 27.45 \\
\multicolumn{1}{l|}{CycleMAE} & 37.54 & 18.01 & \multicolumn{1}{c|}{17.13} & 22.85 & 30.38 & \multicolumn{1}{c|}{22.31} & 24.08 & 24.71 \\
\multicolumn{1}{l|}{BSS/SimCLR} & {\color[HTML]{00009B} \textbf{61.94}} & {\ul 19.58} & \multicolumn{1}{c|}{{\color[HTML]{00009B} \textbf{26.98}}} & {\ul 27.40} & {\ul 31.55} & \multicolumn{1}{c|}{{\ul 41.49}} & {\ul 32.27} & {\ul 34.82} \\
\rowcolor[HTML]{EFEFEF} 
\multicolumn{1}{l|}{\cellcolor[HTML]{EFEFEF}\textbf{MS-UDG}} & {\ul 55.88} & {\color[HTML]{00009B} \textbf{20.03}} & \multicolumn{1}{c|}{\cellcolor[HTML]{EFEFEF}18.65} & {\color[HTML]{00009B} \textbf{32.15}} & {\color[HTML]{00009B} \textbf{39.81}} & \multicolumn{1}{c|}{\cellcolor[HTML]{EFEFEF}{\color[HTML]{00009B} \textbf{42.89}}} & {\color[HTML]{00009B} \textbf{32.45}} & {\color[HTML]{00009B} \textbf{34.92}} \\ \hline
\multicolumn{9}{c}{Label Fraction: 5\%} \\ \hline
\multicolumn{1}{l|}{Source} & \multicolumn{3}{c|}{\textit{painting$\cup$real$\cup$sketch}} & \multicolumn{3}{c|}{\textit{clipart$\cup$infograph$\cup$quichdraw}} &  &  \\
\multicolumn{1}{l|}{Target} & \multicolumn{3}{c|}{\textit{clipart$\cup$infograph$\cup$quichdraw}} & \multicolumn{3}{c|}{\textit{painting$\cup$real$\cup$sketch}} & \textit{\textbf{overall}} & \textit{\textbf{avg.}} \\ \hline
\multicolumn{1}{l|}{BYOL} & 9.60 & 5.09 & \multicolumn{1}{c|}{6.02} & 9.78 & 10.73 & \multicolumn{1}{c|}{3.97} & 7.83 & 7.53 \\
\multicolumn{1}{l|}{MoCo V2} & 28.13 & 13.79 & \multicolumn{1}{c|}{9.67} & 20.8 & 24.91 & \multicolumn{1}{c|}{21.44} & 18.99 & 19.79 \\
\multicolumn{1}{l|}{AdCo} & 30.77 & 18.65 & \multicolumn{1}{c|}{7.75} & 19.97 & 24.31 & \multicolumn{1}{c|}{24.19} & 19.42 & 20.94 \\
\multicolumn{1}{l|}{SimCLR V2} & 34.03 & 17.17 & \multicolumn{1}{c|}{10.88} & 21.35 & 24.34 & \multicolumn{1}{c|}{27.46} & 20.89 & 22.54 \\
\multicolumn{1}{l|}{MAE} & 32.6 & 15.28 & \multicolumn{1}{c|}{13.43} & 24.55 & 30.43 & \multicolumn{1}{c|}{26.07} & 22.88 & 23.73 \\
\multicolumn{1}{l|}{DARLING} & 39.32 & 19.09 & \multicolumn{1}{c|}{10.50} & 21.09 & 30.51 & \multicolumn{1}{c|}{28.49} & 23.31 & 24.83 \\
\multicolumn{1}{l|}{DiMAE} & 42.31 & 18.87 & \multicolumn{1}{c|}{15.00} & 27.02 & 39.92 & \multicolumn{1}{c|}{26.50} & 27.85 & 28.27 \\
\multicolumn{1}{l|}{BrAD} & 37.54 & 18.01 & \multicolumn{1}{c|}{17.13} & 22.85 & 30.38 & \multicolumn{1}{c|}{22.31} & 24.08 & 24.71 \\
\multicolumn{1}{l|}{CycleMAE} & 55.14 & 20.87 & \multicolumn{1}{c|}{19.62} & 27.64 & 40.24 & \multicolumn{1}{c|}{28.71} & 30.80 & 32.04 \\
\multicolumn{1}{l|}{{\color[HTML]{000000} BSS/SimCLR}} & {\color[HTML]{00009B} \textbf{71.21}} & {\color[HTML]{000000} {\ul 20.93}} & \multicolumn{1}{c|}{{\color[HTML]{00009B} \textbf{32.42}}} & {\color[HTML]{000000} {\ul 36.68}} & {\color[HTML]{000000} {\ul 41.49}} & \multicolumn{1}{c|}{{\color[HTML]{00009B} \textbf{52.75}}} & {\color[HTML]{000000} {\ul 39.73}} & {\color[HTML]{000000} {\ul 42.58}} \\
\rowcolor[HTML]{EFEFEF} 
\multicolumn{1}{l|}{\cellcolor[HTML]{EFEFEF}{\color[HTML]{000000} \textbf{MS-UDG}}} & {\color[HTML]{000000} {\ul 71.02}} & {\color[HTML]{00009B} \textbf{28.21}} & \multicolumn{1}{c|}{\cellcolor[HTML]{EFEFEF}{\color[HTML]{000000} {\ul 28.51}}} & {\color[HTML]{00009B} \textbf{40.07}} & {\color[HTML]{00009B} \textbf{47.05}} & \multicolumn{1}{c|}{\cellcolor[HTML]{EFEFEF}{\color[HTML]{000000} {\ul 49.51}}} & {\color[HTML]{00009B} \textbf{41.33}} & {\color[HTML]{00009B} \textbf{44.06}} \\ \hline
\multicolumn{9}{c}{Label Fraction: 10\%} \\ \hline
\multicolumn{1}{l|}{Source} & \multicolumn{3}{c|}{\textit{painting$\cup$real$\cup$sketch}} & \multicolumn{3}{c|}{\textit{clipart$\cup$infograph$\cup$quichdraw}} &  &  \\
\multicolumn{1}{l|}{Target} & \multicolumn{3}{c|}{\textit{clipart$\cup$infograph$\cup$quichdraw}} & \multicolumn{3}{c|}{\textit{painting$\cup$real$\cup$sketch}} & \textit{\textbf{overall}} & \textit{\textbf{avg.}} \\ \hline
\multicolumn{1}{l|}{BYOL} & 14.55 & 8.71 & \multicolumn{1}{c|}{5.92} & 9.50 & 10.38 & \multicolumn{1}{c|}{4.45} & 8.69 & 8.92 \\
\multicolumn{1}{l|}{MoCo V2} & 32.46 & 18.54 & \multicolumn{1}{c|}{8.05} & 25.35 & 29.91 & \multicolumn{1}{c|}{23.71} & 21.87 & 23.00 \\
\multicolumn{1}{l|}{AdCo} & 32.25 & 17.96 & \multicolumn{1}{c|}{11.56} & 23.35 & 29.98 & \multicolumn{1}{c|}{27.57} & 22.79 & 23.78 \\
\multicolumn{1}{l|}{SimCLR V2} & 37.11 & 19.87 & \multicolumn{1}{c|}{12.33} & 24.01 & 30.17 & \multicolumn{1}{c|}{31.58} & 24.28 & 25.84 \\
\multicolumn{1}{l|}{MAE} & 51.86 & 24.81 & \multicolumn{1}{c|}{23.94} & 41.24 & 54.68 & \multicolumn{1}{c|}{39.41} & 38.85 & 39.32 \\
\multicolumn{1}{l|}{DARLING} & 35.15 & 20.88 & \multicolumn{1}{c|}{15.69} & 25.90 & 33.29 & \multicolumn{1}{c|}{30.77} & 26.09 & 26.95 \\
\multicolumn{1}{l|}{DiMAE} & 70.78 & {\ul 38.06} & \multicolumn{1}{c|}{27.39} & 50.73 & \underline{64.89} & \multicolumn{1}{c|}{55.41} & 49.49 & 51.21 \\
\multicolumn{1}{l|}{BrAD} & 68.27 & 26.60 & \multicolumn{1}{c|}{{\ul34.03}} & 31.08 & 38.48 & \multicolumn{1}{c|}{48.17} & 38.74 & 41.10 \\
\multicolumn{1}{l|}{CycleMAE} & {\ul 74.87} & {\color[HTML]{00009B} \textbf{38.42}} & \multicolumn{1}{c|}{28.32} & {\ul 52.81} & {\color[HTML]{00009B} \textbf{67.13}} & \multicolumn{1}{c|}{{\ul 56.37}} & {\ul 50.78} & {\ul 52.98} \\
\multicolumn{1}{l|}{BSS/SimCLR} & 71.95 & 21.27 & \multicolumn{1}{c|}{33.47} & 39.49 & 44.67 & \multicolumn{1}{c|}{55.42} & 41.57 & 44.38 \\
\rowcolor[HTML]{EFEFEF} 
\multicolumn{1}{l|}{\cellcolor[HTML]{EFEFEF}\textbf{MS-UDG}} & {\color[HTML]{00009B} \textbf{79.70}} & 30.01 & \multicolumn{1}{c|}{\cellcolor[HTML]{EFEFEF}{\color[HTML]{00009B} \textbf{40.11}}} & {\color[HTML]{00009B} \textbf{53.73}} & 63.77 & \multicolumn{1}{c|}{\cellcolor[HTML]{EFEFEF}{\color[HTML]{00009B} \textbf{65.82}}} & {\color[HTML]{00009B} \textbf{53.37}} & {\color[HTML]{00009B} \textbf{55.52}} \\ \bottomrule[1pt]
\end{tabular}%
}
\vskip -0.1in
\label{tab:domainnet}
\end{table*}

\subsection{Experimental Results}
We present the experimental results in Tab.~\ref{tab:PACS} (PACS) and Tab.~\ref{tab:domainnet} (DomainNet). Compared to contrastive-based and generative-based SSL methods MoCo V2, SimCLR V2, BYOL, AdCo, and MAE, MS-UDG outperforms in most cases. Among contrastive-based methods, SimCLR v2, which uses multi-view augmented images as self-supervised signals, performs best. 
Compared with other SSL methods, MAE demonstrates better performance on the DomainNet dataset, second only to SimCLR V2 on the PACS dataset. From the results, these SSL methods perform poorly on downstream tasks with unseen domain data, as they ignore domain covariates. 

\begin{wrapfigure}{r}{0.45\linewidth}
\vskip -0.5in
\begin{center}
\centerline{\includegraphics[width=1.0\linewidth]{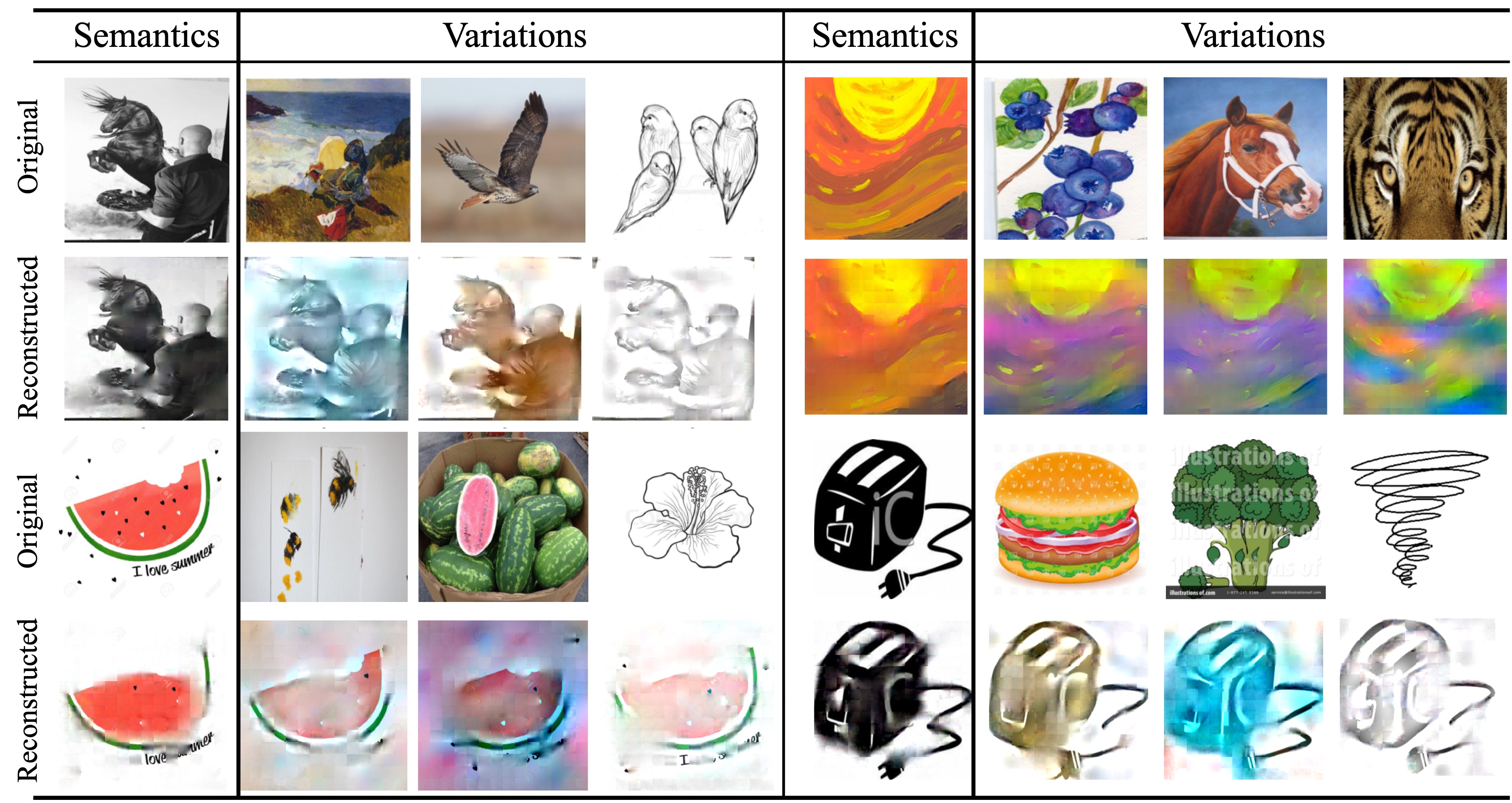}}
\caption{The reconstruction results produced by MS-UDG. Rows 1 and 3 display input images that preserve either semantic content or stylistic variations, derived from four distinct domains within the DomainNet dataset. In contrast, Rows 2 and 4 present reconstructed images using alternative variation representations within the feature space.}
\label{fig:recon}
\end{center}
\vskip -0.35in
\end{wrapfigure}

Compared to the state-of-the-art (SOTA) SSL method SimCLR V2, MS-UDG improves the performance by \textbf{+19.33\%}, \textbf{+22.34\%},\textbf{+26.71\%}, and \textbf{+22.83\%} for label fractions 1\%, 5\%, 10\% and 100\%, respectively, on the PACS dataset. On the DomainNet dataset, MS-UDG improves the SOTA SSL method MAE by \textbf{+17.04\%}, \textbf{+18.45\%}, and \textbf{+14.52\%} in overall accuracy for label fractions 1\%, 5\%, and 10\%. 

From our results, MS-UDG outperforms most contrastive-based and disentanglement-based UDG methods across two datasets. Specifically, in linear evaluation with 1\% and 5\% label fractions, MS-UDG outperforms the SOTA method, BSS, by \textbf{+0.35\%} and \textbf{+6.62\%}, respectively, on the PACS dataset. On the DomainNet dataset, MS-UDG improves by \textbf{+0.18\%} and \textbf{+1.6\%} in linear evaluation compared to BSS. In full fine-tuning with 10\% and 100\% label fractions, MS-UDG surpasses the SOTA method CycleMAE by \textbf{+5.48\%} and \textbf{+1.48\%}, respectively, on the PACS dataset. On the DomainNet dataset, MS-UDG shows an improvement of \textbf{+2.59\%} overall accuracy for the 10\% label fraction.


\subsection{Ablation Study}
\label{ablationstudy}

\begin{wraptable}{r}{0.45\linewidth}
\vskip -0.25in
\caption{Effectiveness of each component of MS-UDG on PACS dataset. All models are trained on 3 selected domains and tested on the remaining domains, which process is repeated for all domains. \textcolor[HTML]{00009B}{\textbf{Best}}  is highlighted.}
\vskip 0.1in
\centering
\resizebox{0.9\linewidth}{!}{%
\begin{tabular}{l|ccc}
\toprule[1pt]
 & \multicolumn{3}{c}{Label Fraction} \\ \cline{2-4} 
\multirow{-2}{*}{Methods} & 1\% & 5\% & 10\% \\ \hline
Baseline ($\mathcal{L}_{suff}$) & 42.45 & 56.17 & 64.81 \\
+$\mathcal{L}_{min}$ & 42.85 & 57.91 & 65.32 \\
+$\mathcal{L}_{max}$ & 44.77 & 57.72 & 67.40 \\
\rowcolor[HTML]{EFEFEF} 
+$\mathcal{L}_{min}$+$\mathcal{L}_{max}$ & {\color[HTML]{00009B} \textbf{45.26}} & {\color[HTML]{00009B} \textbf{63.02}} & {\color[HTML]{00009B} \textbf{68.35}} \\
\toprule[1pt]
\end{tabular}%
}
\vskip -0.1in
\label{tab:ablation}
\end{wraptable}

\textbf{Effectiveness of Each Component of MS-UDG.} Since MS-UDG is based on two optimization objectives, we investigate the impact of each target. Our baseline is the native InfoNCE loss $\mathcal{L}_{suff}$. Next, we introduce the IDM and SROM modules and evaluate $\mathcal{L}_{min}$ and $\mathcal{L}_{max}$, separately. Finally, we assess the performance of the entire method, incorporating $\mathcal{L}_{suff}$, $\mathcal{L}_{min}$, and $\mathcal{L}_{max}$.

We evaluate the effectiveness on the PACS dataset. From Tab.~\ref{tab:ablation}, both applying $\mathcal{L}_{min}$ and  $\mathcal{L}_{max}$ improve the baseline, while the performance can be further improved by combining $\mathcal{L}_{min}$ and $\mathcal{L}_{max}$ by \textbf{+2.81\%}, \textbf{6.85\%}, and \textbf{+3.54\%} under 1\%, 5\%, and 10\% label fractions. Combining the two losses further improves the results, aligning with our theoretical analysis. Minimizing $I(s;v)$ and maximizing $I(v;x|S)$ reduce the semantic-irrelevant information $I(s;x|S)$.


\textbf{Visualization of Reconstruction.} Fig.~\ref{fig:recon} illustrates reconstructed images with consistent semantics and varying styles, showcasing MS-UDG's strong capability to separate features.


\section{Conclusion and Limitation}
\label{conclusion}
In conclusion, we tackle the challenge of semantic disentanglement in UDG without domain labels by formulating it as a constrained optimization problem. Our framework, MS-UDG, integrates contrastive learning with novel modules to effectively separate semantics and variation, achieving state-of-the-art performance. However, MS-UDG is primarily tailored for scenarios with pronounced domain discrepancies. This assumption may limit its effectiveness in settings with subtle covariate shifts, where improvements in disentanglement are less evident.



%% file: neurips_2025.bbl
\begin{thebibliography}{10}

\bibitem{albuquerque2019generalizing}
I.~Albuquerque, J.~Monteiro, M.~Darvishi, T.~H. Falk, and I.~Mitliagkas.
\newblock Generalizing to unseen domains via distribution matching.
\newblock {\em arXiv preprint arXiv:1911.00804}, 2019.

\bibitem{alemi2016deep}
A.~A. Alemi, I.~Fischer, J.~V. Dillon, and K.~Murphy.
\newblock Deep variational information bottleneck.
\newblock {\em arXiv preprint arXiv:1612.00410}, 2016.

\bibitem{ben2010theory}
S.~Ben-David, J.~Blitzer, K.~Crammer, A.~Kulesza, F.~Pereira, and J.~W. Vaughan.
\newblock A theory of learning from different domains.
\newblock {\em Machine learning}, 79:151--175, 2010.

\bibitem{cha2021swad}
J.~Cha, S.~Chun, K.~Lee, H.-C. Cho, S.~Park, Y.~Lee, and S.~Park.
\newblock Swad: Domain generalization by seeking flat minima.
\newblock {\em Advances in Neural Information Processing Systems}, 34:22405--22418, 2021.

\bibitem{chen2020simple}
T.~Chen, S.~Kornblith, M.~Norouzi, and G.~Hinton.
\newblock A simple framework for contrastive learning of visual representations.
\newblock In {\em International conference on machine learning}, pages 1597--1607. PMLR, 2020.

\bibitem{chen2020big}
T.~Chen, S.~Kornblith, K.~Swersky, M.~Norouzi, and G.~E. Hinton.
\newblock Big self-supervised models are strong semi-supervised learners.
\newblock {\em Advances in neural information processing systems}, 33:22243--22255, 2020.

\bibitem{chen2020improved}
X.~Chen, H.~Fan, R.~Girshick, and K.~He.
\newblock Improved baselines with momentum contrastive learning.
\newblock {\em arXiv preprint arXiv:2003.04297}, 2020.

\bibitem{dosovitskiy2020image}
A.~Dosovitskiy.
\newblock An image is worth 16x16 words: Transformers for image recognition at scale.
\newblock {\em arXiv preprint arXiv:2010.11929}, 2020.

\bibitem{Federici2020Learninginfobottleneck}
M.~Federici, A.~Dutta, P.~Forré, N.~Kushman, and Z.~Akata.
\newblock Learning robust representations via multi-view information bottleneck.
\newblock In {\em International Conference on Learning Representations}, 2020.

\bibitem{fukunaga2013introduction}
K.~Fukunaga.
\newblock {\em Introduction to statistical pattern recognition}.
\newblock Elsevier, 2013.

\bibitem{gui2024surveyssl}
J.~Gui, T.~Chen, J.~Zhang, Q.~Cao, Z.~Sun, H.~Luo, and D.~Tao.
\newblock A survey on self-supervised learning: Algorithms, applications, and future trends.
\newblock {\em IEEE Transactions on Pattern Analysis and Machine Intelligence}, 2024.

\bibitem{guo2024domain}
K.~Guo and B.~C. Lovell.
\newblock Domain-aware triplet loss in domain generalization.
\newblock {\em Computer Vision and Image Understanding}, 243:103979, 2024.

\bibitem{harary2022unsupervised}
S.~Harary, E.~Schwartz, A.~Arbelle, P.~Staar, S.~Abu-Hussein, E.~Amrani, R.~Herzig, A.~Alfassy, R.~Giryes, H.~Kuehne, et~al.
\newblock Unsupervised domain generalization by learning a bridge across domains.
\newblock In {\em Proceedings of the IEEE/CVF Conference on Computer Vision and Pattern Recognition}, pages 5280--5290, 2022.

\bibitem{he2022masked}
K.~He, X.~Chen, S.~Xie, Y.~Li, P.~Doll{\'a}r, and R.~Girshick.
\newblock Masked autoencoders are scalable vision learners.
\newblock In {\em Proceedings of the IEEE/CVF conference on computer vision and pattern recognition}, pages 16000--16009, 2022.

\bibitem{he2020momentum}
K.~He, H.~Fan, Y.~Wu, S.~Xie, and R.~Girshick.
\newblock Momentum contrast for unsupervised visual representation learning.
\newblock In {\em Proceedings of the IEEE/CVF conference on computer vision and pattern recognition}, pages 9729--9738, 2020.

\bibitem{hu2021adco}
Q.~Hu, X.~Wang, W.~Hu, and G.-J. Qi.
\newblock Adco: Adversarial contrast for efficient learning of unsupervised representations from self-trained negative adversaries.
\newblock In {\em Proceedings of the IEEE/CVF Conference on Computer Vision and Pattern Recognition}, pages 1074--1083, 2021.

\bibitem{hu2020domain}
S.~Hu, K.~Zhang, Z.~Chen, and L.~Chan.
\newblock Domain generalization via multidomain discriminant analysis.
\newblock In {\em Uncertainty in artificial intelligence}, pages 292--302. PMLR, 2020.

\bibitem{kim2018disentangling}
H.~Kim and A.~Mnih.
\newblock Disentangling by factorising.
\newblock In {\em International conference on machine learning}, pages 2649--2658. PMLR, 2018.

\bibitem{li2017deeper}
D.~Li, Y.~Yang, Y.-Z. Song, and T.~M. Hospedales.
\newblock Deeper, broader and artier domain generalization.
\newblock In {\em Proceedings of the IEEE international conference on computer vision}, pages 5542--5550, 2017.

\bibitem{liang2024factorized}
P.~P. Liang, Z.~Deng, M.~Q. Ma, J.~Y. Zou, L.-P. Morency, and R.~Salakhutdinov.
\newblock Factorized contrastive learning: Going beyond multi-view redundancy.
\newblock {\em Advances in Neural Information Processing Systems}, 36, 2024.

\bibitem{liu2025medmap}
T.~Liu, Z.~Tan, M.~Chen, X.~Yang, H.~Jiang, and K.~Huang.
\newblock Medmap: Promoting incomplete multi-modal brain tumor segmentation with alignment.
\newblock {\em IEEE Journal of Biomedical and Health Informatics}, 2025.

\bibitem{liu2023towards}
Y.~Liu, Y.~Chen, M.~Gou, C.-T. Huang, Y.~Wang, W.~Dai, and H.~Xiong.
\newblock Towards unsupervised domain generalization for face anti-spoofing.
\newblock In {\em Proceedings of the IEEE/CVF International Conference on Computer Vision}, pages 20654--20664, 2023.

\bibitem{liu2023promoting}
Y.~Liu, Y.~Wang, Y.~Chen, W.~Dai, C.~Li, J.~Zou, and H.~Xiong.
\newblock Promoting semantic connectivity: Dual nearest neighbors contrastive learning for unsupervised domain generalization.
\newblock In {\em Proceedings of the IEEE/CVF Conference on Computer Vision and Pattern Recognition}, pages 3510--3519, 2023.

\bibitem{mitrovic2021representation}
J.~Mitrovic, B.~McWilliams, J.~C. Walker, L.~H. Buesing, and C.~Blundell.
\newblock Representation learning via invariant causal mechanisms.
\newblock In {\em International Conference on Learning Representations}, 2021.

\bibitem{nam2021reducingsagnet}
H.~Nam, H.~Lee, J.~Park, W.~Yoon, and D.~Yoo.
\newblock Reducing domain gap by reducing style bias.
\newblock In {\em Proceedings of the IEEE/CVF Conference on Computer Vision and Pattern Recognition}, pages 8690--8699, 2021.

\bibitem{nam2021reducing}
H.~Nam, H.~Lee, J.~Park, W.~Yoon, and D.~Yoo.
\newblock Reducing domain gap by reducing style bias.
\newblock In {\em Proceedings of the IEEE/CVF Conference on Computer Vision and Pattern Recognition}, pages 8690--8699, 2021.

\bibitem{niu2015multimvdg}
L.~Niu, W.~Li, and D.~Xu.
\newblock Multi-view domain generalization for visual recognition.
\newblock In {\em Proceedings of the IEEE international conference on computer vision}, pages 4193--4201, 2015.

\bibitem{oord2018representation}
A.~v.~d. Oord, Y.~Li, and O.~Vinyals.
\newblock Representation learning with contrastive predictive coding.
\newblock {\em arXiv preprint arXiv:1807.03748}, 2018.

\bibitem{oquab2023dinov2}
M.~Oquab, T.~Darcet, T.~Moutakanni, H.~Vo, M.~Szafraniec, V.~Khalidov, P.~Fernandez, D.~Haziza, F.~Massa, A.~El-Nouby, et~al.
\newblock Dinov2: Learning robust visual features without supervision.
\newblock {\em arXiv preprint arXiv:2304.07193}, 2023.

\bibitem{pan2025structure}
T.~Pan, Z.~Tan, K.~Guo, D.~Xu, W.~Xu, C.~Jiang, X.~Guo, Y.~Qi, and Y.~Cheng.
\newblock Structure-aware semantic discrepancy and consistency for 3d medical image self-supervised learning.
\newblock {\em arXiv preprint arXiv:2507.02581}, 2025.

\bibitem{peng2019moment}
X.~Peng, Q.~Bai, X.~Xia, Z.~Huang, K.~Saenko, and B.~Wang.
\newblock Moment matching for multi-source domain adaptation.
\newblock In {\em Proceedings of the IEEE/CVF international conference on computer vision}, pages 1406--1415, 2019.

\bibitem{scalbert2023towards}
M.~Scalbert, M.~Vakalopoulou, and F.~Couzini{\'e}-Devy.
\newblock Towards domain-invariant self-supervised learning with batch styles standardization.
\newblock {\em arXiv preprint arXiv:2303.06088}, 2023.

\bibitem{scalberttowards}
M.~Scalbert, M.~Vakalopoulou, and F.~Couzinie-Devy.
\newblock Towards domain-invariant self-supervised learning with batch styles standardization.
\newblock In {\em The Twelfth International Conference on Learning Representations}, 2024.

\bibitem{tan2024personalize}
Z.~Tan, X.~Yang, T.~Pan, T.~Liu, C.~Jiang, X.~Guo, Q.~Wang, A.~Nguyen, Y.~Qi, K.~Huang, et~al.
\newblock Personalize to generalize: Towards a universal medical multi-modality generalization through personalization.
\newblock {\em arXiv preprint arXiv:2411.06106}, 2024.

\bibitem{tan2024interpret}
Z.~Tan, X.~Yang, Q.~Wang, A.~Nguyen, and K.~Huang.
\newblock Interpret your decision: Logical reasoning regularization for generalization in visual classification.
\newblock {\em Advances in Neural Information Processing Systems}, 37:18166--18204, 2024.

\bibitem{tishby2000informationbottleneck}
N.~Tishby, F.~C. Pereira, and W.~Bialek.
\newblock The information bottleneck method.
\newblock {\em arXiv preprint physics/0004057}, 2000.

\bibitem{tsai2021selfsupervisedminimal}
Y.-H.~H. Tsai, Y.~Wu, R.~Salakhutdinov, and L.-P. Morency.
\newblock Self-supervised learning from a multi-view perspective.
\newblock In {\em International Conference on Learning Representations}, 2021.

\bibitem{wang2022rethinkingminimal}
H.~Wang, X.~Guo, Z.-H. Deng, and Y.~Lu.
\newblock Rethinking minimal sufficient representation in contrastive learning.
\newblock In {\em Proceedings of the IEEE/CVF Conference on Computer Vision and Pattern Recognition}, pages 16041--16050, 2022.

\bibitem{wang2022generalizingdgsurvey}
J.~Wang, C.~Lan, C.~Liu, Y.~Ouyang, T.~Qin, W.~Lu, Y.~Chen, W.~Zeng, and P.~S. Yu.
\newblock Generalizing to unseen domains: A survey on domain generalization.
\newblock {\em IEEE transactions on knowledge and data engineering}, 35(8):8052--8072, 2022.

\bibitem{wang2020understanding}
T.~Wang and P.~Isola.
\newblock Understanding contrastive representation learning through alignment and uniformity on the hypersphere.
\newblock In {\em International conference on machine learning}, pages 9929--9939. PMLR, 2020.

\bibitem{wang2024disentangled}
X.~Wang, H.~Chen, Z.~Wu, W.~Zhu, et~al.
\newblock Disentangled representation learning.
\newblock {\em IEEE Transactions on Pattern Analysis and Machine Intelligence}, 2024.

\bibitem{xu2021fourier}
Q.~Xu, R.~Zhang, Y.~Zhang, Y.~Wang, and Q.~Tian.
\newblock A fourier-based framework for domain generalization.
\newblock In {\em Proceedings of the IEEE/CVF conference on computer vision and pattern recognition}, pages 14383--14392, 2021.

\bibitem{yang2022cycle}
H.~Yang, X.~Li, S.~Tang, F.~Zhu, Y.~Wang, M.~Chen, L.~Bai, R.~Zhao, and W.~Ouyang.
\newblock Cycle-consistent masked autoencoder for unsupervised domain generalization.
\newblock In {\em The Eleventh International Conference on Learning Representations}, 2022.

\bibitem{yang2022domain}
H.~Yang, S.~Tang, M.~Chen, Y.~Wang, F.~Zhu, L.~Bai, R.~Zhao, and W.~Ouyang.
\newblock Domain invariant masked autoencoders for self-supervised learning from multi-domains.
\newblock In {\em European Conference on Computer Vision}, pages 151--168. Springer, 2022.

\bibitem{yu2024insure}
X.~Yu, H.-H. Tseng, S.~Yoo, H.~Ling, and Y.~Lin.
\newblock Insure: an information theory inspired disentanglement and purification model for domain generalization.
\newblock {\em IEEE Transactions on Image Processing}, 2024.

\bibitem{zhang2025disentangling}
A.~Zhang, H.~Wang, X.~Wang, and T.-S. Chua.
\newblock Disentangling masked autoencoders for unsupervised domain generalization.
\newblock In {\em European Conference on Computer Vision}, pages 126--151. Springer, 2025.

\bibitem{zhang2022towardsdisentagle}
H.~Zhang, Y.-F. Zhang, W.~Liu, A.~Weller, B.~Sch{\"o}lkopf, and E.~P. Xing.
\newblock Towards principled disentanglement for domain generalization.
\newblock In {\em Proceedings of the IEEE/CVF conference on computer vision and pattern recognition}, pages 8024--8034, 2022.

\bibitem{zhang2022towards}
X.~Zhang, L.~Zhou, R.~Xu, P.~Cui, Z.~Shen, and H.~Liu.
\newblock Towards unsupervised domain generalization.
\newblock In {\em Proceedings of the IEEE/CVF Conference on Computer Vision and Pattern Recognition}, pages 4910--4920, 2022.

\bibitem{zhou2022domain}
K.~Zhou, Z.~Liu, Y.~Qiao, T.~Xiang, and C.~C. Loy.
\newblock Domain generalization: A survey.
\newblock {\em IEEE Transactions on Pattern Analysis and Machine Intelligence}, 45(4):4396--4415, 2022.

\end{thebibliography}
